\documentclass[a4paper]{llncs}

\usepackage{graphicx}
\usepackage{subfigure}

\usepackage{amsmath,amsfonts,amssymb}   

\newenvironment{rcases}
  {\left.\begin{aligned}}
  {\end{aligned}\right\rbrace}

\usepackage{tabularx,supertabular,booktabs,lscape,array,longtable}

\usepackage[vlined,linesnumbered,ruled]{algorithm2e}

\usepackage{todonotes} 

\usepackage{url}
\def\none{} 

\usepackage{soul}
\usepackage[normalem]{ulem}

\def\cplex{{\tt IBM-ILOG CPLEX Cplex 12.7}}
\def\unit{{\tt UNIT}}

\begin{document}

\title{Deep Neural Networks as 0-1 Mixed Integer Linear Programs: A Feasibility Study}

\author{Matteo Fischetti$^{*}$ \and Jason Jo$^{+}$}

\institute{$^{*}$ Department of Information Engineering (DEI), University of Padova \email{matteo.fischetti@unipd.it} (corresponding author)\\
$^{+}$ Montreal Institute for Learning Algorithms (MILA) and Institute for Data Valorization (IVADO), Montreal \email{jason.jo.research@gmail.com}}
\maketitle

\medskip
\noindent
\centerline{(submitted to an international conference)} 

\begin{abstract}    
Deep Neural Networks (DNNs) are very popular these days, and are the subject of a very intense investigation. A DNN is made by layers of internal units (or neurons), each of which computes an affine combination of the output of the units in the previous layer, applies a nonlinear operator, and outputs the corresponding value (also known as activation). A commonly-used nonlinear operator is the so-called rectified linear unit (ReLU), whose output is just the maximum between its input value and zero. In this (and other similar cases like max pooling, where the max operation involves more than one input value), one can model the DNN as a 0-1 Mixed Integer Linear Program (0-1 MILP) where the continuous variables correspond to the output values of each unit, and a binary variable is associated with each ReLU to model its yes/no nature. In this paper we discuss the peculiarity of this kind of 0-1 MILP models, and describe an effective bound-tightening technique intended to ease its solution. We also present possible applications of the 0-1 MILP model arising in feature visualization and in the construction of adversarial examples. Preliminary computational results are reported, aimed at investigating (on small DNNs) the computational performance of a state-of-the-art MILP solver when applied to a known test case, namely, hand-written digit recognition.

\end{abstract}


\section{Introduction}\label{sec:intro} 

In this work we address a Deep Neural Network (DNN) with rectified linear units (ReLUs) \cite{relu} or max/average pooling activations. The main contributions of our work can be outlined as follows:

\begin{itemize}
\item[$\bullet$] We investigate a 0-1 Mixed-Integer Linear Programming (0-1 MILP) model of the DNN. We recently became aware that a similar model has been independently proposed in \cite{serra2017} and applied for counting the linear regions of a given DNN. \none{Therefore we cannot claim the model is new. However, to the best of our knowledge the applications and discussions we report in the present paper are original and hopefully of interest.} 
\item[$\bullet$] We discuss the peculiarities of the 0-1 MILP formulation, thus motivating other researchers to derive improved (possibly heuristic) solution schemes exploiting them.   
\item[$\bullet$] We present a bound-tightening mechanism that has a relevant impact in reducing solution times. 
\item[$\bullet$] We discuss two new applications of the 0-1 MILP model in the context of Feature Visualization \cite{bengio} and Adversarial Machine Learning \cite{adv}, the latter being very well suited for our approach as finding (almost) optimal solutions is very important.
\end{itemize}   

\section{A 0-1 MILP model}\label{sec:model} 

Let the DNN be made by $K+1$ (say) layers, numbered from 0 to $K$. Layer 0 is fictitious and corresponds to the input of the DNN, while the last layer, $K$ corresponds to its outputs. Each layer $k\in \{0,1,\dots, K\}$ is made by $n_k$ units (i.e., nodes in networks, or neurons), numbered for 1 to $n_k$. We denote by $\unit(j,k)$ the $j$-th unit of layer $k$.

Let $x^k \in \Re^{n_k}$ be the output vector of layer $k$, i.e., $x^k_j$ is the output of $\unit(j,k)$ ($j=1,\dots,n_k$). As already mentioned, layer 0 corresponds to the DNN input, hence $x_j^0$ is the $j$-th input value (out of $n_0$) for the DNN. Analogously, $x^K_j$ is the $j$-th output value (out of $n_K$) of the DNN viewed as a whole. For each layer $k \ge 1$, $\unit(j,k)$ computes its output vector $x^k$ through the formula $$x^k = \sigma(W^{k-1} x^{k-1} + b^{k-1}),$$ where $\sigma(\cdot)$ is a nonlinear function (possibly depending on $j$ and $k$), and $W^{k-1}$ (resp. $b^{k-1}$) is a \emph{given} matrix of weights (resp., vector of biases). 

As in many applications, we will assume that $\sigma$ is a \emph{rectified linear unit}, i.e., the equations governing the DNN are
\begin{align} 
	x^k = ReLU(W^{k-1} x^{k-1} + b^{k-1}), \ \ \ \ &k=1,\dots,K  \label{eq:DNN}
\end{align}                           
where, for a real vector $y$, $ReLU(y) := \max\{0, y\}$ (componentwise).
           
Note that the weight/bias matrices $(W,b)$ can (actually, have to) contain negative entries, while all the output vectors $x^k$ are nonnegative, with the possible exception of the vector $x^0$ that represents the input of the DNN as a whole.

To get a valid 0-1 MILP model for a given DNN, one needs to study the basic scalar equation $$x = ReLU(w^T y + b).$$ To this end, one can write the linear conditions

\begin{align}                                         
	w^T y + b = x - s, \ \ \  x \ge 0, \ \ \ s \ge 0   \label{eq:basic}
\end{align}
to decouple the positive and negative part of the ReLU input. Unfortunately, the solution $(x, s)$ of constraints \eqref{eq:basic} is not unique (as it should be because ReLU() is in fact a function), because one can always take any scalar $\delta \ge 0$ and obtain a still-feasible solution $(x + \delta, s + \delta)$. Imposing $\delta=0$ is therefore the only critical issue to be addressed when modeling the $ReLU$ operator. (Note that minimizing the sum $x + s$ is not a viable option here, as this would alter the DNN nature and will tend to reduce the absolute value of the ReLU input). 

To impose that at least one of the two terms $x$ and $s$ must be zero, one could add to \eqref{eq:basic} the bilinear (complementary) condition $x\, s \le 0$, which is equivalent to $x \, s = 0$ as both terms in the product are required to be nonnegative. 

A second option (which is the one we applied in our study) is to introduce a binary \emph{activation} variable $z$ and to impose the logical implications
\begin{align}
\begin{rcases}  
z = 1 \rightarrow x \le 0  \ \ \ \ \\
z = 0 \rightarrow s \le 0    \ \ \ \ \\   
z \in \{0,1\}   \ \ \ \            
\end{rcases} \label{eq:zeta}    
\end{align}  
The above ``indicator constraints'' are accepted as such by modern MILP solvers, and are internally converted into proper linear inequalities of the type $x \le M^+ (1-z_j)$ and $s \le M^- z_j$ (assuming that finite nonnegative values $M^+$ and $M^-$ can be computed such that $-M^- \le w^T y + b \le M^+$) and/or are handed implicitly by the solution algorithm. 

Using a binary activation variable $z^k_j$ for each $\unit(j,k)$ then leads to the following 0-1 MILP formulation of the DNN:  
\def\mipmodel{\eqref{eq:model1}--\eqref{eq:model4}}                                               
\begin{align}
\min \sum_{k=0}^K \sum_{j=1}^{n_k} c_j^k x_j^k + \sum_{k=1}^K \sum_{j=1}^{n_k} \gamma_j^k z_j^k \ \ \ \ \ \ \ \ \ \ \ \ \ \ \ \ \ \ \ \ \ \ \ \ \ \ \ \ \ &  \label{eq:model1} \\
\begin{rcases} 
\sum_{i=1}^{n_{k-1}}w_{ij}^{k-1} x_i^{k-1} + b_j^{k-1} = x_j^k - s_j^k \ \ \ \ \\
x_j^k, s_j^k \ge 0 \ \ \ \   \\
z_j^k \in \{0,1\} \ \ \ \ \\
z_j^k = 1 \rightarrow x_j^k \le 0 \ \ \ \ \\
z_j^k = 0 \rightarrow s_j^k \le 0 \ \ \ \ \\
\end{rcases}  
k=1,\dots,K, \ j=1,\dots,n_k \label{eq:model2} \\
lb_j^0 \le x_j^0 \le ub_j^0,  \ \ \ \ \ \ j=1,\dots,n_0 \ \  \ \ \ \ \ \ \ \ \ \ \ \ \ \ \ \ \ \ \ \ \ \ \ \  \label{eq:model3}\\
\begin{rcases} 
	lb_j^k \le x_j^k \le ub_j^k \ \ \ \  \\
\overline{lb}_j^k \le s_j^k \le  \overline{ub}_j^k \ \ \ \  \\
\end{rcases}
k=1,\dots,K, \ j=1,\dots,n_k.  \ \ \ \ \ \ \label{eq:model4} 
\end{align}                                                    
In the above formulation, all weights $w_{ij}^{k-1}$ and biases $b_j^k$ are given (constant) parameters; the same holds for the objective function costs $c_j^k$ and $\gamma_j^k$, that can be defined according to the specific problem at hand (some relevant cases will be addressed in the next section). Conditions \eqref{eq:model2} define the ReLU output for each unit, while \eqref{eq:model3}--\eqref{eq:model4} impose known lower and upper bounds on the $x$ and $s$ variables: for $k=0$, these bounds apply to the DNN input values $x_j^0$ and depend on their physical meaning, while for $k\ge 1$ one has $lb_j^k=\overline{lb}_j^k = 0$ and $ub_j^k, \, \overline{ub}_j^k \in \Re_+ \cup \{+\infty\}$. 

\none{
Besides ReLU activations, some modern DNNs such as Convolutional Neural Networks (CNNs) \cite{mnist,cnn2} involve multi-input units that perform the following \emph{average/max pooling} operations:  
\begin{align} 
	x &= AvgPool(y_1,\dots,y_t) =  \frac{1}{t} \ \sum_{i=1}^t y_i \label{eq:avg} \\
	x &= MaxPool(y_1,\dots,y_t) = \max\{y_1, \dots, y_t\}.   \label{eq:max} 
\end{align}                                                            
The first operation \eqref{eq:avg} is just linear and can trivially be incorporated in our MILP model, while \eqref{eq:max} can be expressed by introducing a set of binary variables $z_1,\cdots,z_t$ (that represent $\arg\max$) along with the following constraints:        

\begin{align}
\sum_{i=1}^t z_i = 1  \ \ \ \ \ \ \ \ \ \ \ \ \ \ \ \ \ \ \ \ \ \ \ \ \label{eq:max1} \\ 
\begin{rcases}
x \ge y_i,  \ \ \ \  \\
z_i = 1 \rightarrow x \le y_i \ \ \ \ \   \\   
z_i \in \{0,1\} \ \ \ \ \      
\end{rcases}   
i=1,\cdots,t   \label{eq:max2}
\end{align}

It should be noticed that indicator constraints such as those appearing in \eqref{eq:zeta} or \eqref{eq:max2}, as well their bilinear equivalent like $x_j^k \ s_j^k \le 0$, tend to produce very hard mixed-integer instances that challenge the current state-of-the-art solvers. As a matter of fact, the evaluation of the practical feasibility of model \mipmodel\ was one of the main motivations of our work.
}

\paragraph{Discussion} Here are some comments about the 0-1 MILP model above.
\begin{itemize}
\item[1.]	If one fixes the input $x^0$ of the DNN (i.e., if one sets $lb^0_j=ub^0_j$ for all $j=1,\dots,n_0$), then all the other $x$ variables in the model are fixed to a unique possible value---the one that the DNN itself would compute by just applying equations \eqref{eq:DNN} by increasing layers. As to the binary $z$ variables, they are also defined uniquely, with the only possible ``degenerate'' exception of the binary variable $z_j^k$ corresponding to a ReLU unit that receives a zero input, hence its actual value is immaterial. 

\item[2.] Because of the above, the 0-1 MILP model \mipmodel\ cannot be infeasible, and actually \emph{any} (possibly random) input vector $x^0$ satisfying the bounds conditions \eqref{eq:model3} can easily be extended (in a unique way) to produce a feasible solution. (Note that the same does not hold if one randomly fixes the activation binary variables $z_j^k$.) This is in contrast with other 0-1 MILP models with indicator (or big-M) constraints, for which even finding a feasible solution is a hard task. In this view, powerful refinement heuristics such a \emph{local branching} \cite{locbra}, \emph{polishing} \cite{polish}, or \emph{proximity search} \cite{proxy} can be used to improve a given (possibly random) solution. This is important in practice as it suggests a hybrid solution scheme in which initial heuristic solutions are found through fast methods such as gradient descent, and then refined using MILP technology.

\item[3.] It is known \cite{indicator} that, for 0-1 MILP models like \mipmodel, the definition of tight upper bounds for the continuous  variables appearing in the indicator constraints plays a crucial role for the practical resolution of the model itself. Modern MILP solvers are able to automatically define reasonable such upper bounds, propagating the lower/upper bounds on the input layer $0$ to the other ones. However, these upper bounds can be rather inaccurate. We found that a much better option, very much in the spirit of \cite{indicator}, is instead as follows: We scan the units by increasing layers $k=1,\dots,K$. For the current $\unit(j,k)$, we remove from the model all the constraints (and variables) related to all the other units in the same layer or in the next ones, and solve twice the resulting model: one to maximize $x_j^k$ and the other to maximize $s_j^k$. The resulting optimal values (or their optimistic estimate returned by the MILP solver after a short time limit) can then be used to define a tight bound on the two variables $x_j^k$ and $s_j^k$, respectively. Due to the acyclic nature of the DNN, the tightened bounds computed in one iteration can be used in the next ones, i.e., the method always solves a 0-1 MILP with very tight upper bounds on the continuous variables. In the end, the computed tightened bounds can be saved in a file, so they can be exploited for any future optimization of the same DNN. \none{Note that the upper bound computation for each layer can be distributed (without communication) on a cluster of parallel computers, thus reducing preprocessing time.}
	
\end{itemize}
                      
\section{Applications}\label{sec:applications}   

\none{
Model \mipmodel\ is (un)fortunately \emph{not} suited for training. In training, indeed, one has a number of training examples, each associated with a different input $x^0$. So, in this setting, $x^0$ can be considered to be given, while the variables to optimize are the weights $w_j^{k}$ and biases $b_j^k$. It then follows that, for any internal layer $k\ge 2$, the $x$'s still play the role of  variables (as they depend on the weights/biases of the previous layer), so the terms $w_{ij}^{k-1} x_i^{k-1}$ are in fact bilinear. 
}

On the other hand, previous experiences of using MILP technology for training (such as the one reported in \cite{svm}) seem to indicate that, due to overfitting, insisting on finding proven optimal solutions is not at all a clever policy for training.
   
Instead, our model is more suited for applications where the weights are fixed, and one wants to compute the best-possible input example according to some linear objective function, as in the relevant cases discussed below. In those cases, indeed, overfitting the given DNN is actually a desired property.

\subsection{Experimental setup}\label{sec:experiments} 
We considered a very standard classification problem: hand-written digit recognition of an input 28 x 28 figure. Each of the 784 pixels of the figure is normalized to become a gray-level in $[0,1]$, where 1 means white and 0 black. 
\none{
As in \cite{serra2017}, the MNIST \cite{mnist} dataset was used to train a simple DNN with 3 hidden layers with (8, 8, 8) ReLUs, plus a last layer made by 10 units to classify digits ``0'' to ``9'', reaching (after 50 epochs) an accuracy of 93.04\% on the test set.  

} 
   
\subsection{Feature Visualization}\label{sec:features} 

\begin{figure}
\begin{center}
\includegraphics[width=.45\textwidth]{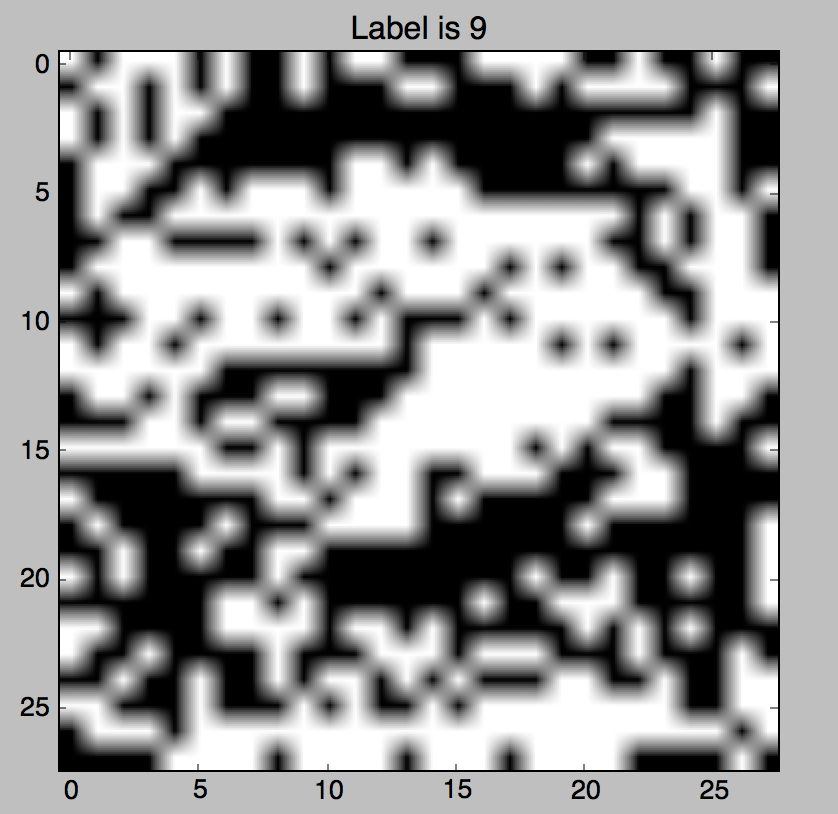}
\includegraphics[width=.435\textwidth]{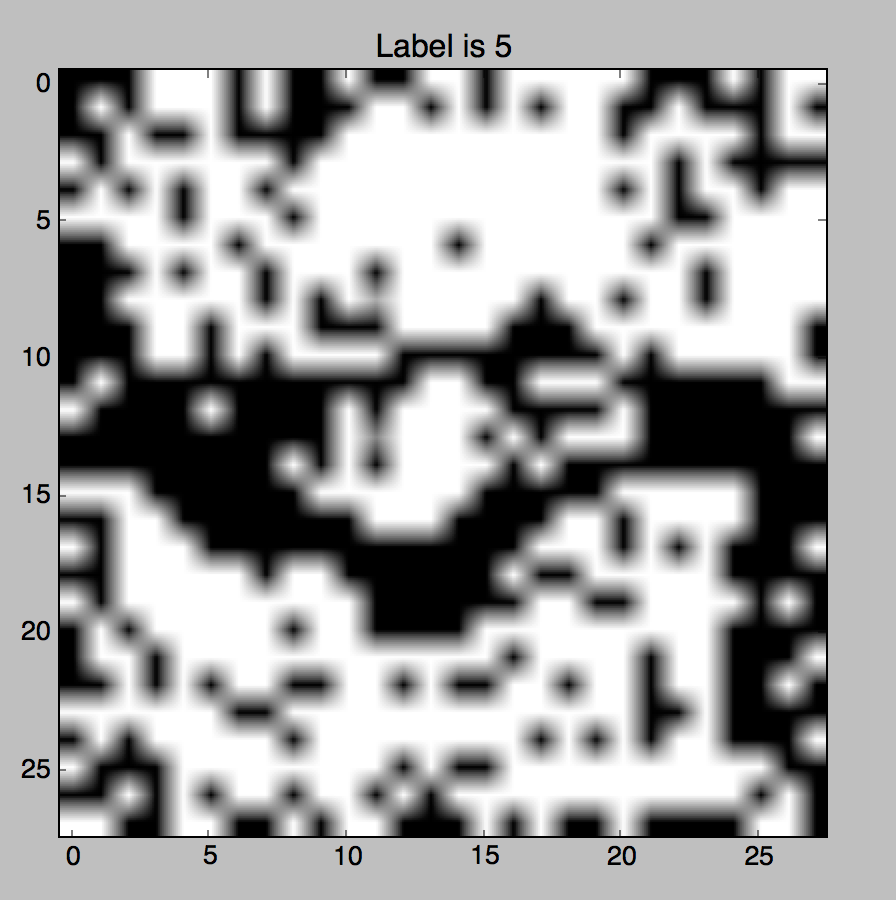}

\caption{Input examples maximizing the activation of some hidden units; no visual pattern can be identified in these provably optimal solutions.} \label{fig:fv}
\end{center}
\end{figure}

Following \cite{bengio} we used our 0-1 MILP model to find input examples that maximize the activation $x_j^k$ of each unit $\unit(j,k)$. For our simple DNN, each of the resulting models could be solved within 1 second (and very often much faster) when using, on the fly, the bound strengthening procedure described in the previous section (the solver was more than one order of magnitude slower without this feature).  Some max-activating input examples are depicted in Figure~\ref{fig:fv}, and show that no nice visual pattern can be recognized (at least, for our DNN).

\none{
It should be noticed that, before our work, the computation of the max activations was performed in the literature by using a greedy ascent method that can be trapped by local optimal solutions. According to our experience with a preliminary implementation of a gradient-based method, many times one is even unable to find any meaningful solution with activation strictly larger than zero. In our view, the capability of finding provable optimal solutions (and, in any case, very good solutions)is definitely a strength of our approach.}

\subsection{Building Adversarial Examples}\label{sec:adversarial}    

In our view, this is the most intriguing application of the MILP technology, due to its ability to compute (almost) optimal solutions that are intended to point out some hidden weaknesses of the DNN of interest. The problem here is to slightly modify a given DNN input so that to produce a wrong output. The construction of these optimized ``adversarial examples'' is the core of adversarial machine learning \cite{adv}.

\begin{figure}
\begin{center}
\includegraphics[width=.45\textwidth]{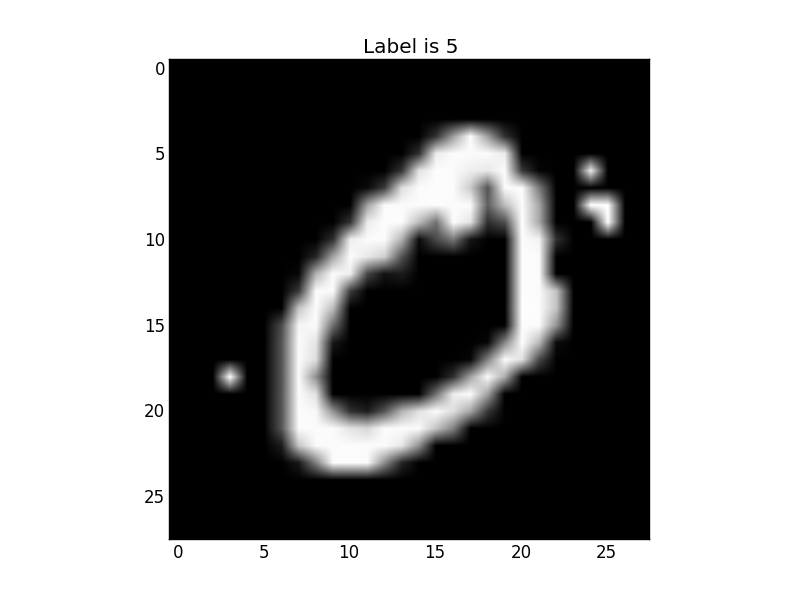}
\includegraphics[width=.45\textwidth]{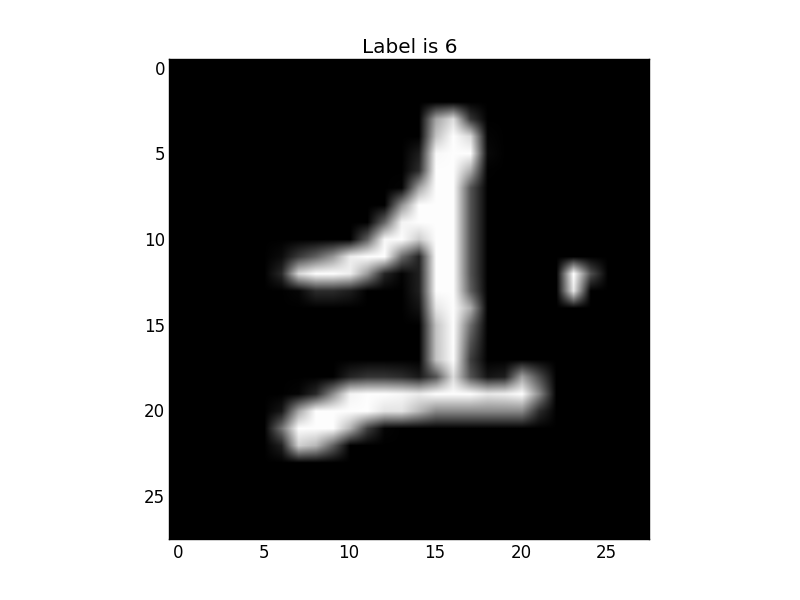}
\includegraphics[width=.45\textwidth]{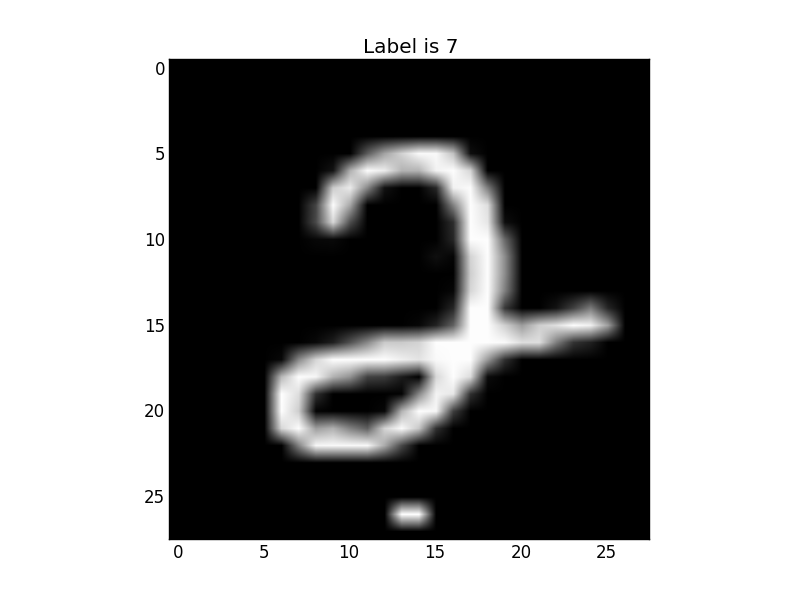}
\includegraphics[width=.45\textwidth]{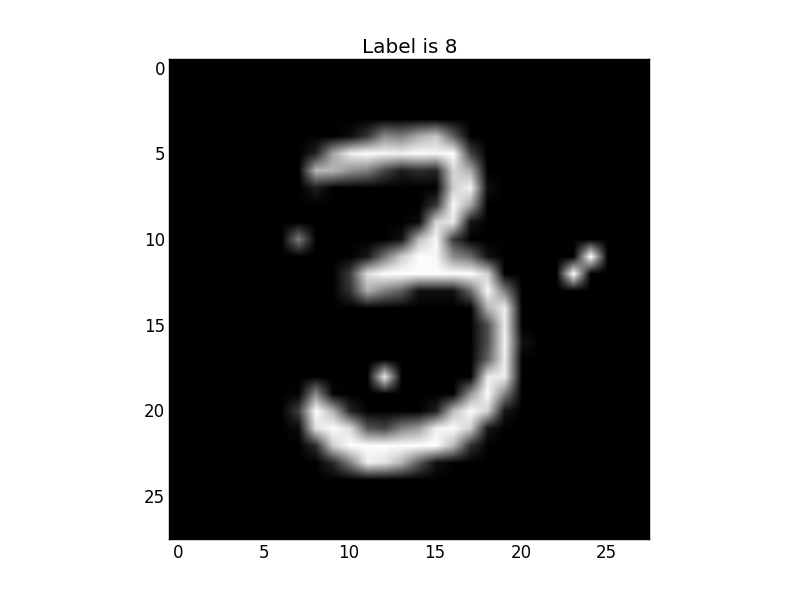}
\includegraphics[width=.45\textwidth]{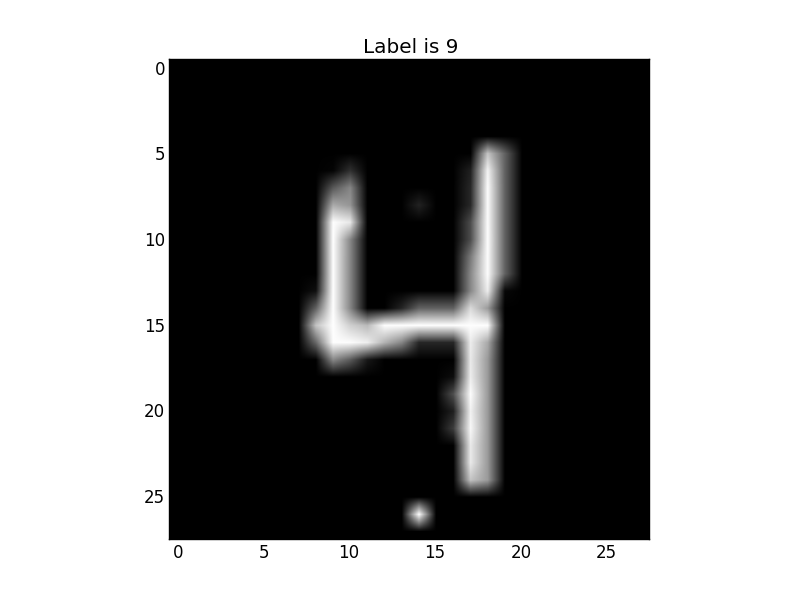}
\includegraphics[width=.45\textwidth]{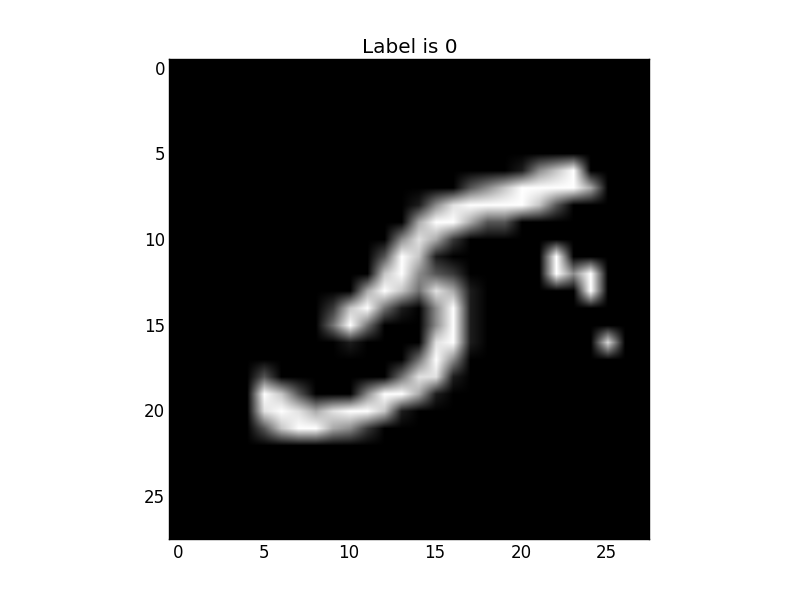}
\includegraphics[width=.45\textwidth]{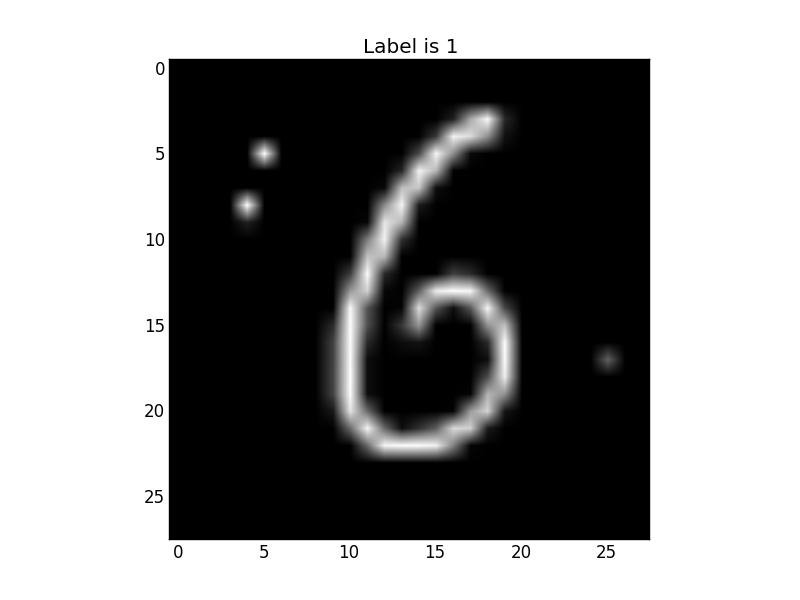}
\includegraphics[width=.45\textwidth]{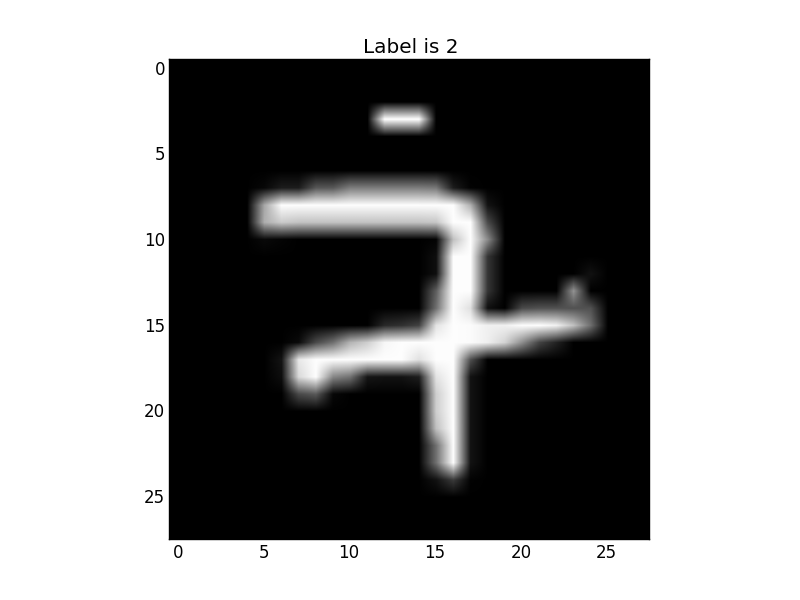}
\includegraphics[width=.45\textwidth]{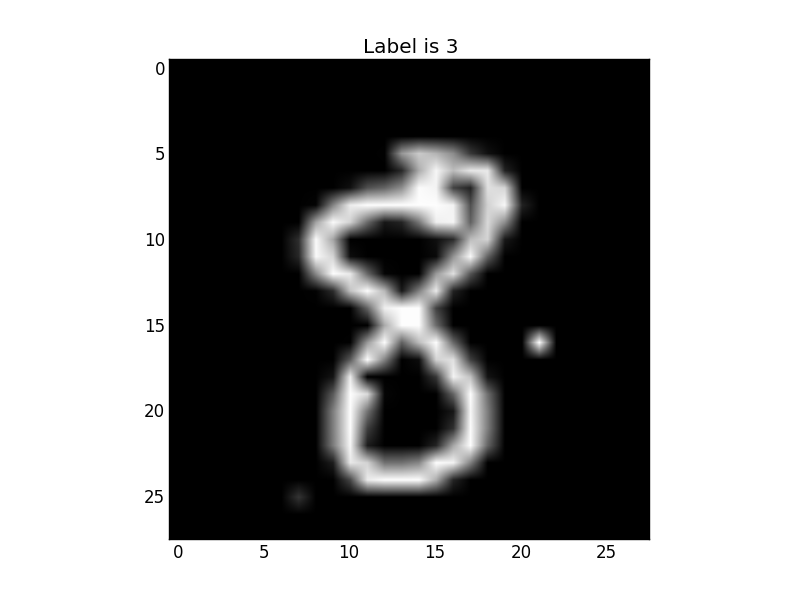}
\includegraphics[width=.45\textwidth]{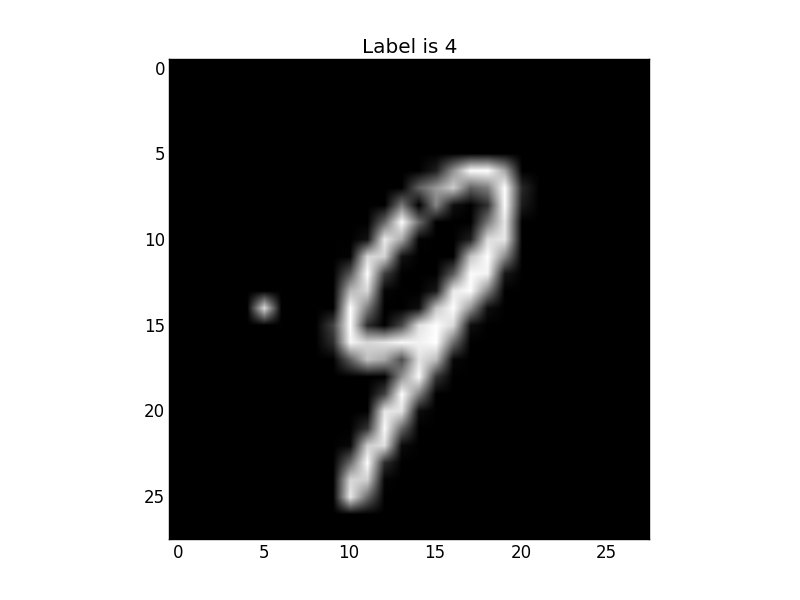}

\caption{Adversarial examples computed through our 0-1 MILP model; the reported label is the one having maximum activation according to the DNN \none{(that we imposed to be the true label plus 5, modulo 10).} Note that the change of just few well-chosen pixels often suffices to fool the DNN and to produce a wrong classification.} \label{fig:adv}
\end{center}
\end{figure}

In our setting, we are given an input figure $\tilde x^0$ which is classified correctly by our DNN as a certain digit $\tilde d$ (say), and we want to produce a similar figure $x^0$ which is wrongly classified as $d\not = \tilde d$. In our experiments, we set $d = (\tilde d + 5) \mod 10$. E.g., we require that a ``0'' must be classified as ``5'', and a ``6'' as a ``1''; see Figure~\ref{fig:adv} for an illustration. To this end we impose, in the final layer, that the activation of the required (wrong) digit is at least 20\% larger than any other activations. Due to the MILP flexibility, this just requires adding to \mipmodel\ the linear conditions 
\begin{align}
x^K_{d+1} \ge 1.2\, x_{j+1}^K, \ \ \ j \in \{0,\dots,9\} \setminus \{d\}. \label{eq:adv1} 
\end{align}
In order to reduce the $L_1$-norm distance between $x^0$ and $\tilde x^0$, we minimize the ad-hoc objective function $\sum_{j=1}^{n_0} d_j$, where the additional continuous variables $d_j$ must satisfy the following linear inequalities to be added to model \mipmodel: 
\begin{align}   
-d_j \le x_j^0  - \tilde x_j^0 \le d_j, \  d_j \ge 0, \ \ \ \ \  \ \hbox{for~} j=1,\dots,n_0. \label{eq:adv2} 
\end{align}
Figure~\ref{fig:adv} illustrates the power of the approach, in that the model is able to locate 2-3 critical pixels whose change tricks the DNN.  

\begin{figure}
\begin{center}
\includegraphics[width=.45\textwidth]{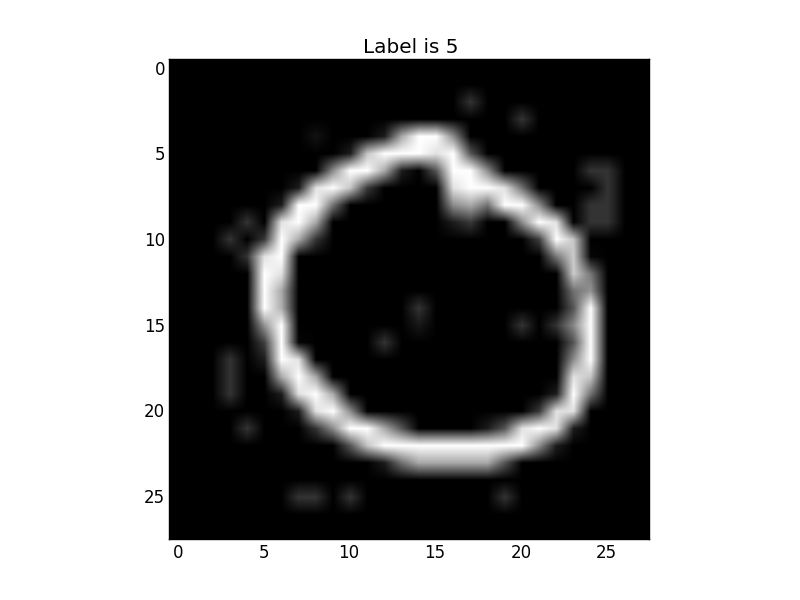}
\includegraphics[width=.45\textwidth]{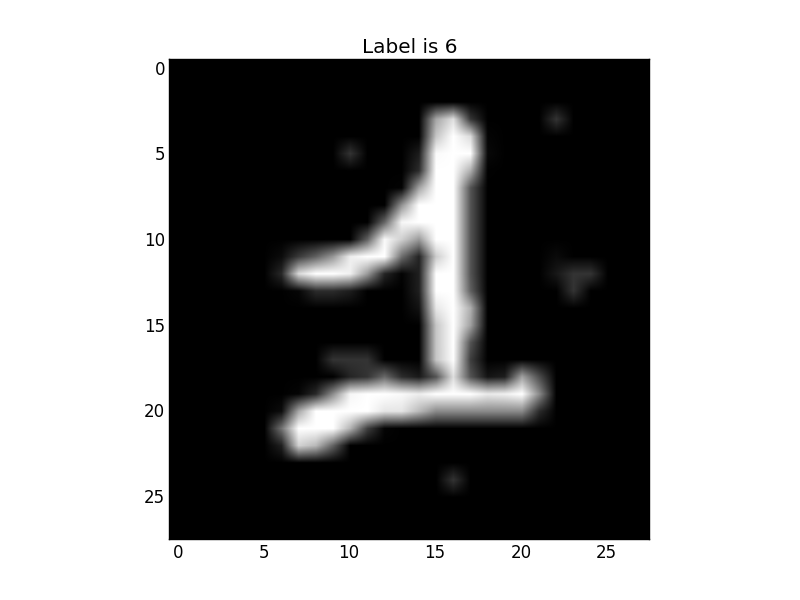}
\includegraphics[width=.45\textwidth]{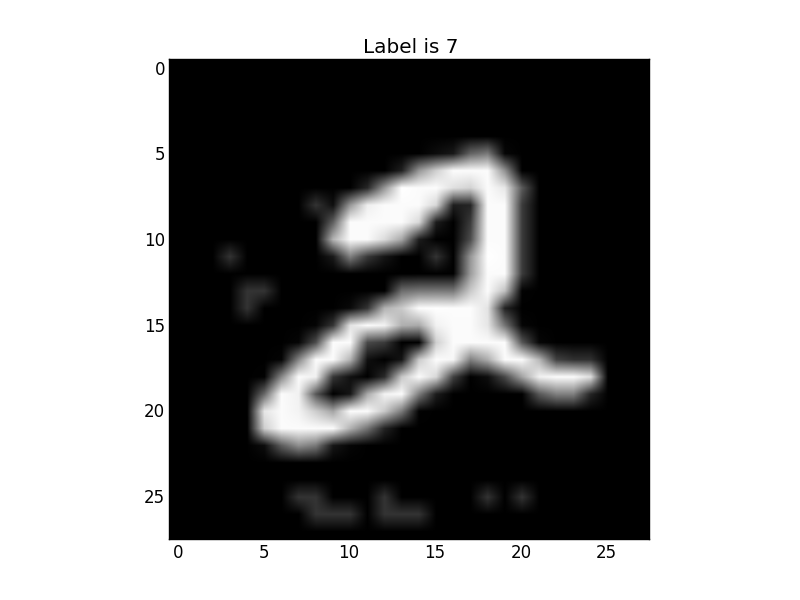}
\includegraphics[width=.45\textwidth]{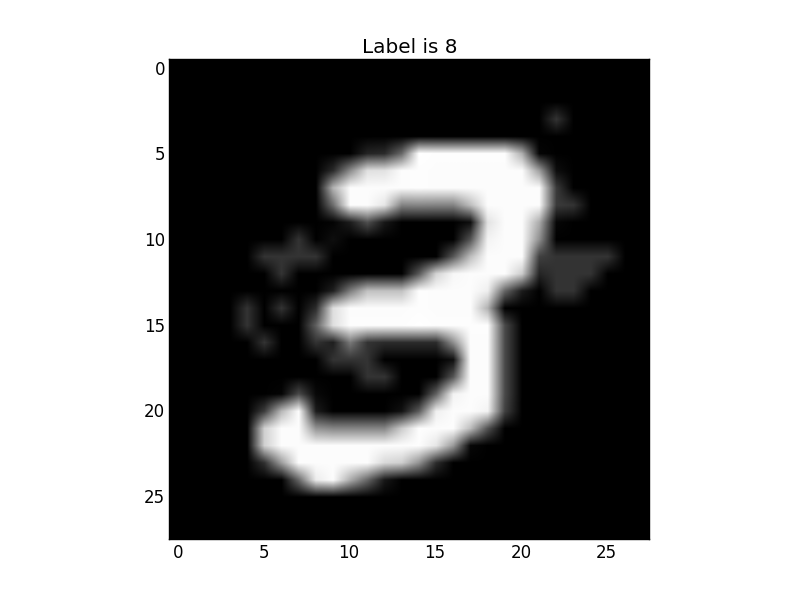}
\includegraphics[width=.45\textwidth]{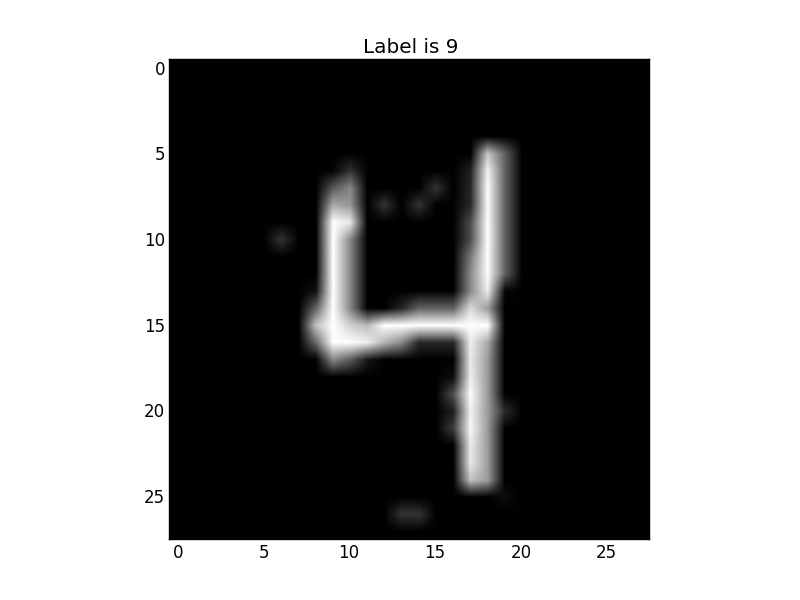}
\includegraphics[width=.45\textwidth]{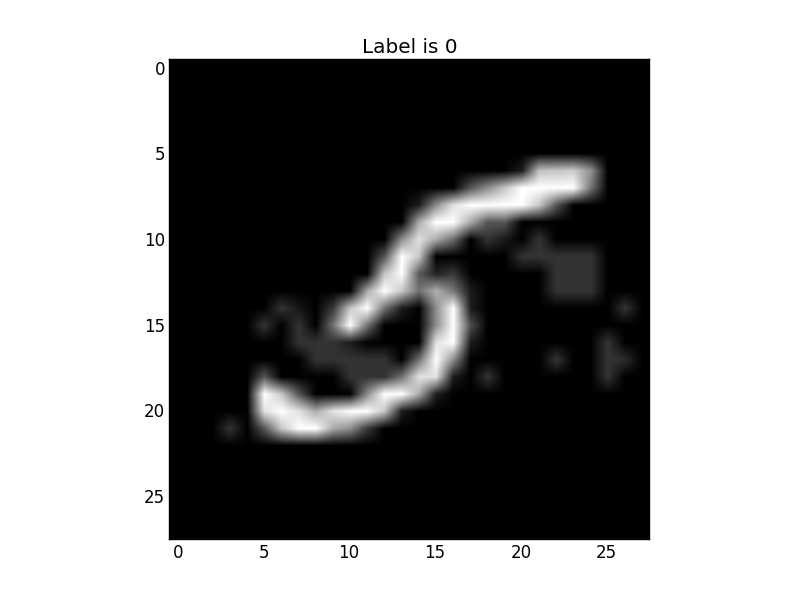}
\includegraphics[width=.45\textwidth]{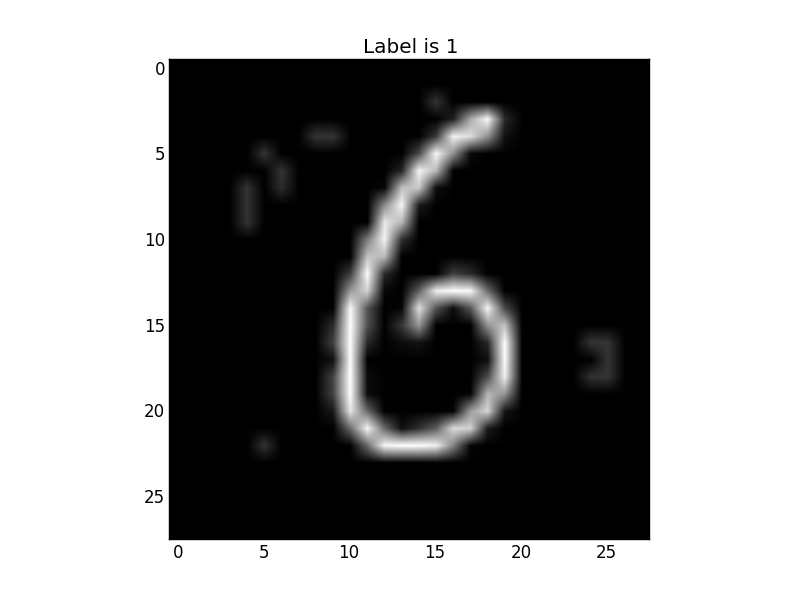}
\includegraphics[width=.45\textwidth]{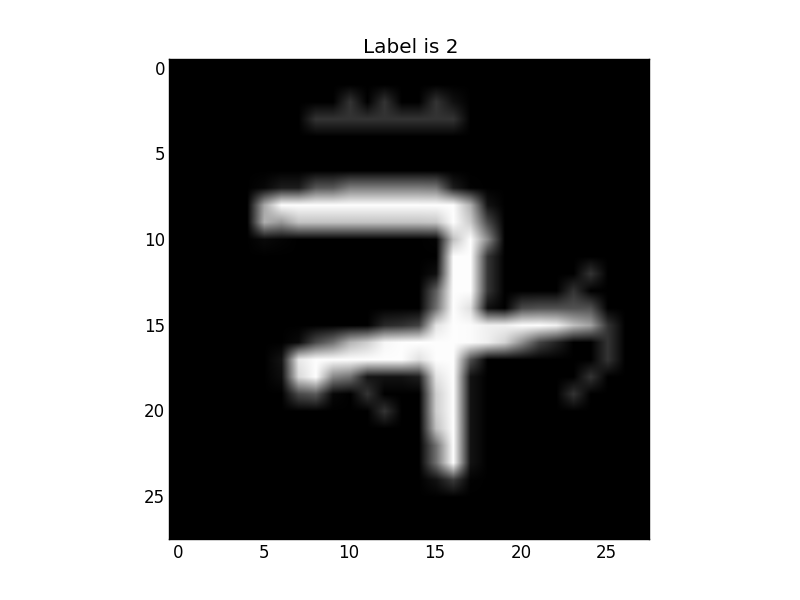}
\includegraphics[width=.45\textwidth]{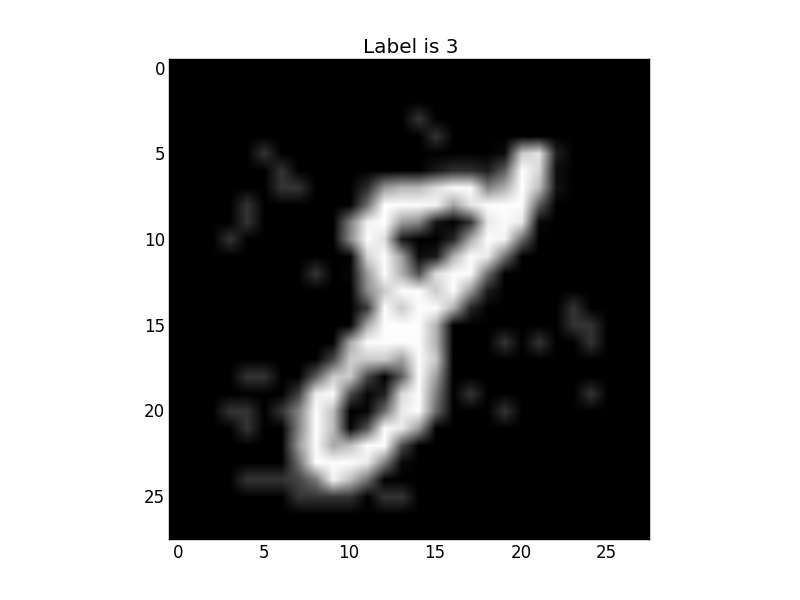}
\includegraphics[width=.45\textwidth]{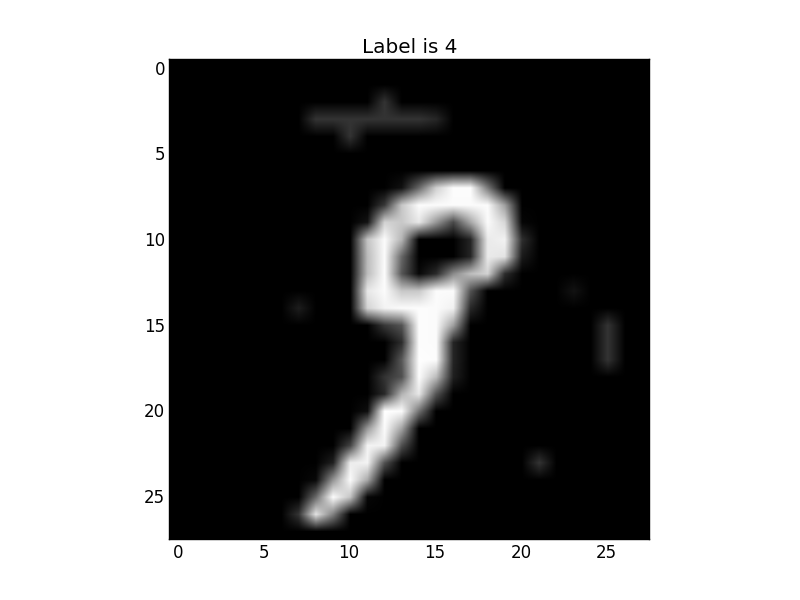}
\caption{\none{Adversarial examples computed through our 0-1 MILP model as in Figure~\ref{fig:adv}, but imposing that the no pixel can be changed by more than 0.2 (through the additional conditions $d_j \le 0.2$ for all $j$).}} \label{fig:adv2}
\end{center}
\end{figure}

\begin{figure}
\begin{center}
\includegraphics[width=.45\textwidth]{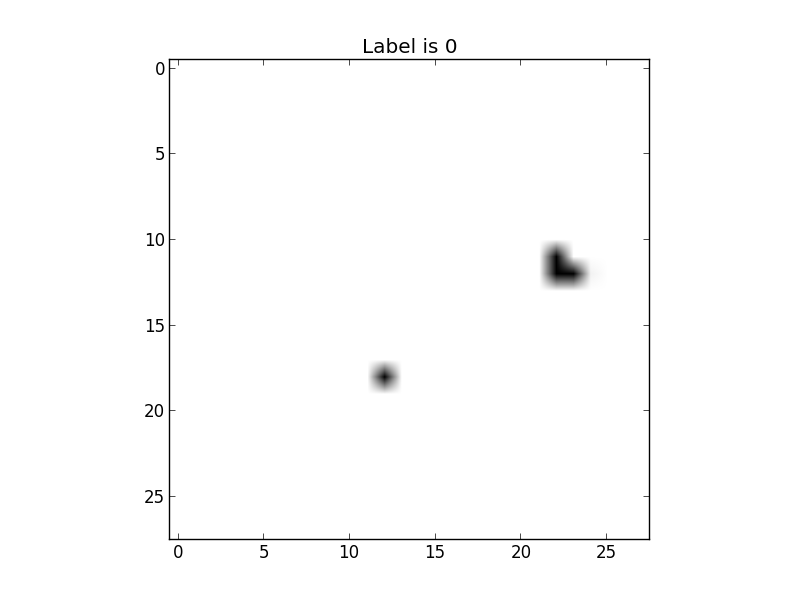}
\includegraphics[width=.45\textwidth]{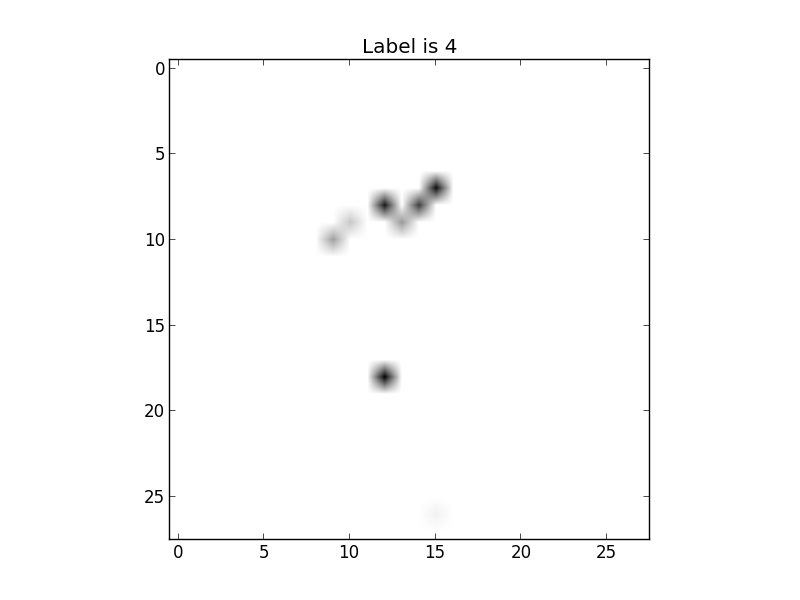}
\includegraphics[width=.45\textwidth]{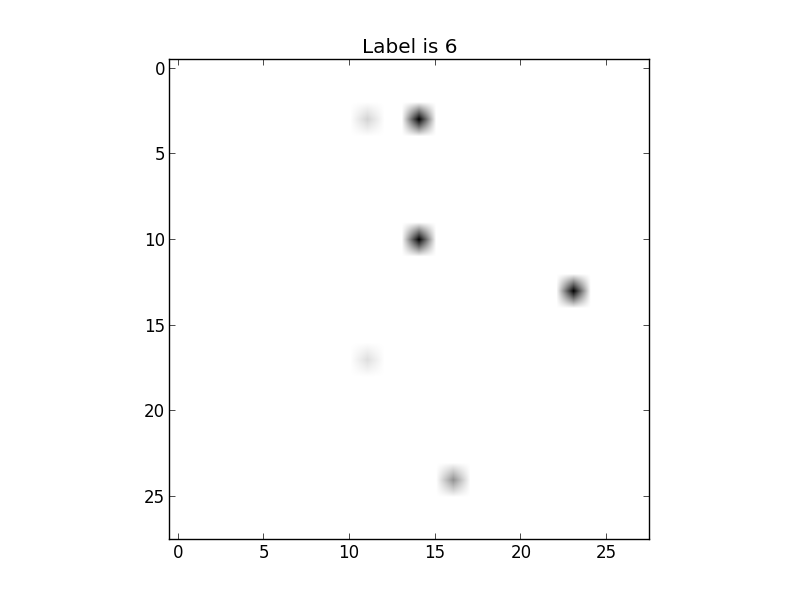}
\includegraphics[width=.45\textwidth]{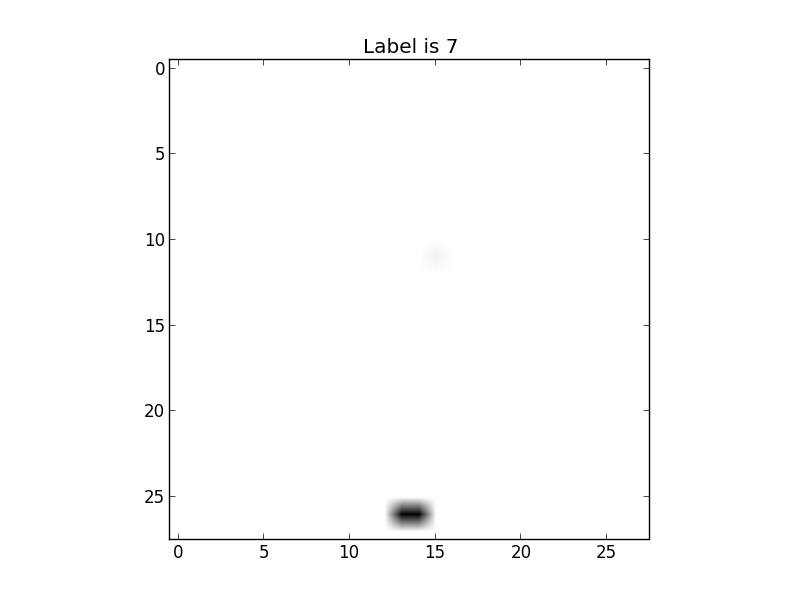}
\includegraphics[width=.45\textwidth]{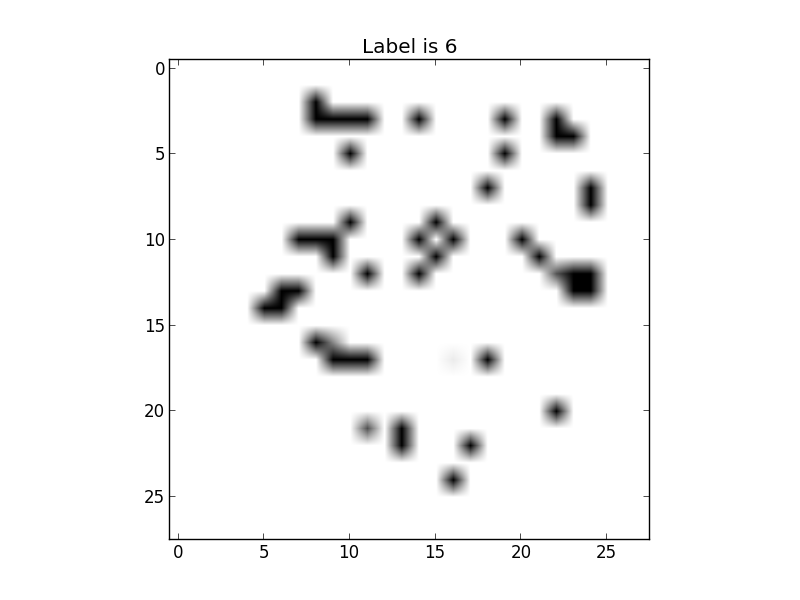}
\includegraphics[width=.45\textwidth]{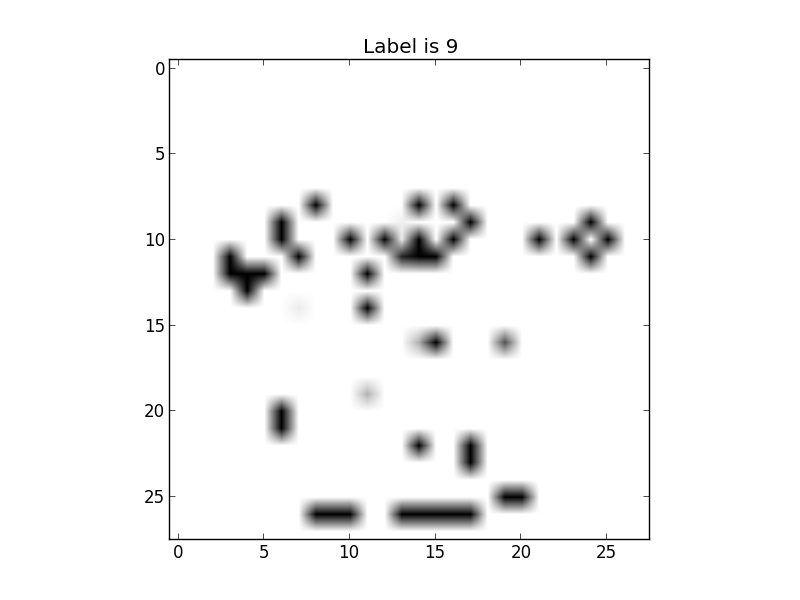}
\includegraphics[width=.45\textwidth]{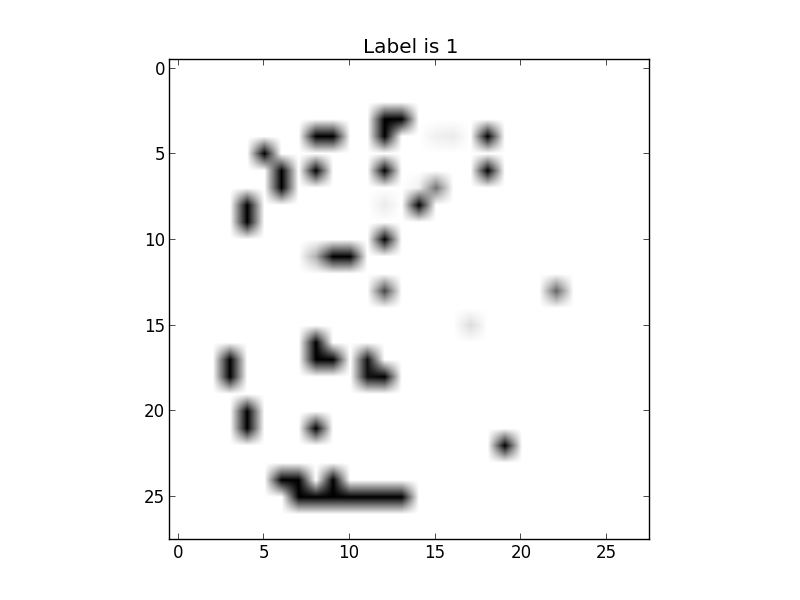}
\includegraphics[width=.45\textwidth]{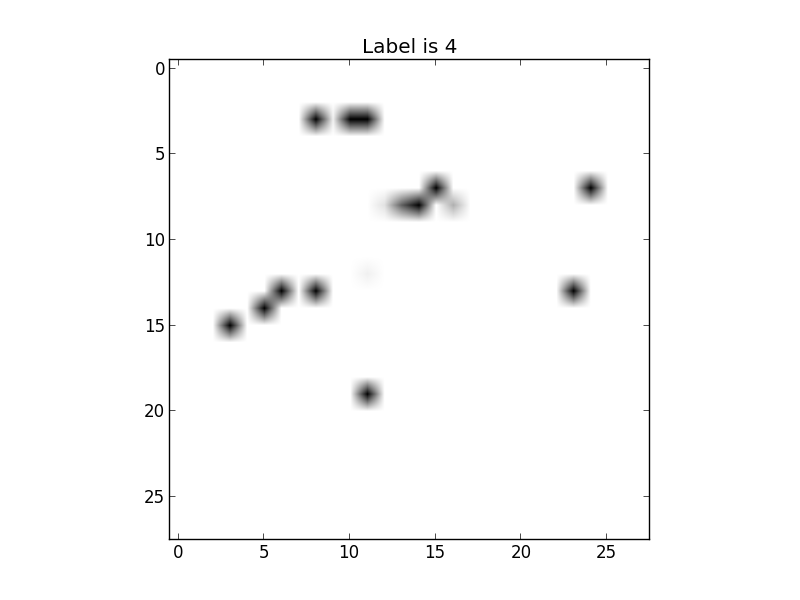}
\caption{\none{Pixel changes (absolute value) that suffice to trick the DNN: the four top subfigures correspond to the model where pixels can change arbitrarily, while those on the bottom refer to the case where each pixel cannot change by more than 0.2 (hence more pixels need be changed). To improve readability, the black/white map has been reverted, i.e., white corresponds to unchanged pixels (i.e., to $d_j=0$).}} \label{fig:adv3}
\end{center}
\end{figure}

\none{A key advantage of our method over previous approaches is that one can easily impose constraints like the one requiring that the final activation of the wrong label is at least 20\% larger than any other activation. Similar constraints can be imposed to the input figure $x^0$, requiring e.g. a maximum number of changed pixel in the input figure, or a maximum deviation of each pixel with respect to the $\tilde x^0$; see Figure~\ref{fig:adv2} for an illustration. This control is an important novelty with respect to other adversarial example generation methods, and it can hopefully allow for a more qualitative probing of what exactly a DNN has learned. In addition, as Figure~\ref{fig:adv3} clearly shows, the optimized solutions are very different form the random noise used by the standard methods typically used in this kind of experiments.}

\section{Computational performance}\label{sec:computational}

\none{
This section is aimed at investigating the practical performance of a state-of-the-art MILP solver (\cplex\ in our case) to construct adversarial examples for not-too-small DNNs. We used the same experimental MNIST setup as in the previous section, and addressed DNNs with the following structure:  
\begin{itemize} 
	\item{\tt DNN1:} 8+8+8 internal units in 3 hidden layers, as in \cite{serra2017};
	\item{\tt DNN2:} 8+8+8+8+8+8 internal units in 6 hidden layers;  
	\item{\tt DNN3:} 20+10+8+8 internal units in 4 hidden layers;  
	\item{\tt DNN4:} 20+10+8+8+8 internal units in 5 hidden layers;  
	\item{\tt DNN5:} 20+20+10+10+10 internal units in 5 hidden layers.  
\end{itemize}                 
All DNNs involve an additional input layer (i.e., layer 0) with 784 entries for the pixels of the 28x28 input figure, and an additional output layer (i.e., layer $K$) with 10 units to classify the ten digits.

All DNNs were trained for 50 epochs and produced a test-set (top-1) accuracy of 93-96\%. The best weights/biases were used to build our basic model \mipmodel, that was then modified for the adversarial case by adding the distance variables $d_j$'s and the associated constraints \eqref{eq:adv1}--\eqref{eq:adv2}. All $d_j$ variables have no upper bound, meaning that we do not impose any limit to the change of the input pixels; see Figure~\ref{fig:adv} for an illustration of the typical adversarial examples computed through this model.

Before running the final experiments, a preprocessing phase was applied (for each DNNs) to tighten the variable bounds as described in Section~\ref{sec:model}, item 3, with a time limit of 5 minutes for each bound computation (this time limit was reached just few times for the largest DNNs). The tightened bounds for all the $x_j^k$ and $s_j^k$ variables were saved in a file and used in the runs reported in the present section under the label ``improved model''.
                                                        
Table~\ref{tab:one} reports some statistics of our runs (average values over 100 runs for each DNN and each model). 
Computational times refer to the use of the state-of-the-art MILP solver \cplex\ \cite{cplex} on a standard 4-core notebook equipped with an Intel i7 @ 2.3GHz processor and 16 GB RAM---the GPU being not used by the MILP solver. A time limit of 300 sec.s was imposed for each run. 

In the table, column ``\%solved'' reports the percentage of the instances that have been solved to proven optimality within the time  limit, while columns ``nodes'' and ``time (s)'' give, respectively, the average number of branching nodes and of computing time (in CPU seconds) over all instances; time-limit instances count as 300 sec.s. Finally, column ``\%gap'' gives the percentage gap between the best upper bound and the best lower bound computed for the instance at hand (instances solved to proven optimality having a gap of zero).                                                                                                  

According to the table, the basic model gets into trouble for our three largest DNNs, as it was not able to solve to prove optimality a large percentage of the instances and returned a significant gap in the end. On the other hand, the improved model consistently outperforms the basic one, and starts having difficulties only with our most-difficult network (DNN5). The difference in terms of computing time and number of branching nodes is also striking.    

} 

\begin{table}[htbp]
\tabcolsep=4pt
\begin{center}
\begin{tabular}{l|rrrr|rrrr}
\hline
\multicolumn{1}{c}{} &
\multicolumn{4}{|c}{basic model} &
\multicolumn{4}{|c}{improved model} \\
 & {\%solved} & {\%gap } & nodes & {time (s)} &{\%solved} & {\%gap } & nodes & {time (s)} \\   
\hline                                                    
{\tt DNN1}  &  100 &  0.0 & 1,903   &   1.0   & 100 &  0.0 &  552     & 0.6    \\    
{\tt DNN2}  &  97  &  0.2 & 77,878  &  48.2   & 100 &  0.0 &  11,851  & 7.5    \\    
{\tt DNN3}  &  64  & 11.6 & 228,632 & 158.5   & 100 &  0.0 &  20,309  & 12.1   \\    
{\tt DNN4}  &  24  & 38.1 & 282,694 & 263.0   & 98  &  0.7 &  68,563  & 43.9   \\    
{\tt DNN5}  &  7   & 71.8 & 193,725 & 290.9   & 67  & 11.4 &  76,714  & 171.1 \\     

\hline
\end{tabular}
\end{center}
\caption{\none{Comparison of the two basic and improved models with a time limit of 300 CPU sec.s, clearly showing the importance of bound tightening in the improved model.} }
\label{tab:one}
\end{table}


\section{Conclusions and future work}\label{sec:conclusions} 

We have addressed a 0-1 Mixed-Integer Linear model for Deep Neural Networks with ReLUs and max/average pooling. This is a very first step in the direction of using discrete optimization as a core tool in the study of neural networks.

We have discussed the specificities of the proposed model, and we have described an effective bound-tightening technique to significantly reduce solution times. Although the model is not suited for training (as it becomes bilinear in this setting), it can be useful to construct optimized input examples for a given (already trained) neural network. In this spirit, we have reported its application to two relevant problems in Machine Learning such as Feature Visualization and Adversarial Machine Learning. In particular, the latter qualifies as as a natural setting for mixed-integer optimization, in that one calls for (almost) optimal solutions that fool the neural network by ``overfitting'' it.


For small DNNs, our model can be solved to proven optimality in a matter of seconds on a standard notebook (no GPU needed). However, for larger and more realistic DNNs the computing time can become too large. For example, even in the MNIST setting, DNNs of size (30, 20, 10, 10, 10, 8, 8, 8) or (50, 50, 50, 20, 8, 8) lead to computing times of one hour or more. In those hard cases, one should resort to heuristics possibly based on a restricted version of the model itself, in the vein of \cite{locbra,polish,proxy}.                                          
       
Future work should therefore address the reduction of the computational effort involved in the exact solution of the model, as well as new heuristic methods for building adversarial examples for large DNNs. Finding new deep learning applications of our mixed-integer model is also an interesting topic for future research.

\section*{Acknowledgements}   
The research of the first author was partially funded by the Vienna Science and Technology Fund (WWTF) through project ICT15-014, any by MiUR, Italy, through project PRIN2015 ``Nonlinear and Combinatorial Aspects of Complex Networks''. \none{The research of the second author was funded by the Institute for Data Valorization (IVADO), Montreal.} We thank Yoshua Bengio and Andrea Lodi for helpful discussions.

\bibliography{dnn}
\bibliographystyle{plain}

\end{document}